\documentclass{article}

\usepackage{microtype}
\usepackage{graphicx}
\usepackage{booktabs}

\usepackage{amsmath}
\allowdisplaybreaks
\usepackage{amssymb}
\usepackage{mathtools}
\usepackage{amsthm}

\usepackage{hyperref}

\usepackage[preprint]{icml2026}

\theoremstyle{plain}
\newtheorem{theorem}{Theorem}[section]

\newtheorem{lemma}[theorem]{Lemma}

\theoremstyle{definition}

\theoremstyle{remark}

\usepackage{amsfonts}       
\usepackage{nicefrac} 
\usepackage{physics}
\usepackage{comment}

\usepackage[textwidth=2.1cm,disable]{todonotes}
\newcommand{\bigO}{\mathcal{O}}
\renewcommand{\textapprox}{\raisebox{0.5ex}{\texttildelow}}

\newcommand{\btheta}{\boldsymbol{\theta}}

\newcommand{\EX}{\mathbb{E}}

\newcommand{\pmsize}{\normalsize}

\makeatletter

\newcounter{phase}[algorithm]
\newlength{\phaserulewidth}
\newcommand{\setphaserulewidth}{\setlength{\phaserulewidth}}

\makeatother

\setphaserulewidth{.7pt}

\icmltitlerunning{\textsc{LayUp}: Asynchronous decentralized gradient descent with {\sc lay}er-wise {\sc up}dates}

\begin{document}

\twocolumn[
\icmltitle{\textsc{LayUp}: Asynchronous decentralized gradient descent with {\sc lay}er-wise {\sc up}dates}



\icmlsetsymbol{equal}{*}

\begin{icmlauthorlist}
\icmlauthor{Cabrel Teguemne Fokam}{rub}
\icmlauthor{Marcel Nieveler}{tud,ubi}
\icmlauthor{Lukas König}{citec,rub}
\icmlauthor{Khaleelulla Khan Nazeer}{tud}
\icmlauthor{David Kappel}{citec}
\icmlauthor{Anand Subramoney}{rhul}
\end{icmlauthorlist}

\icmlaffiliation{citec}{Center for Cognitive Interaction Technology, Universität Bielefeld, Germany}
\icmlaffiliation{rub}{Ruhr Universität Bochum, Germany}
\icmlaffiliation{ubi}{Bielefeld University}
\icmlaffiliation{tud}{Chair of Highly-Parallel VLSI Systems and Neuro-Microelectronics, Technische Universität Dresden, Germany}
\icmlaffiliation{rhul}{Department of Computer Science, Royal Holloway, University of London, United Kingdom}

\icmlcorrespondingauthor{Cabrel Teguemne Fokam}{cabrel.teguemnefokam@ini.ruhr-uni-bochum.de}
\icmlkeywords{Machine Learning, ICML}

]



\printAffiliationsAndNotice{} 

\setcounter{footnote}{1}


\begin{abstract}
The increasing size of deep learning models has made distributed training across multiple devices essential.
Synchronous, centralized methods incur large communication and synchronization overheads.
Communication efficient algorithms can reduce these overheads, but often require extra buffers, remain sensitive to stragglers or parameter drift.
We present LayUp, an asynchronous decentralized SGD method with layer-wise updates.
LayUp asynchronously exchanges incremental layer-wise updates during backpropagation.
It uses randomized gossip communication, enabling updates to be applied as soon as they are available without buffering.
These design choices reduce parameter drift and improve robustness to stragglers.
We establish a theoretical upper bound for the gradient bias introduced by layer-wise updates and prove convergence of LayUp.
We empirically validate LayUp on vision and language modeling tasks, showing convergence up to \textapprox 32\% faster in terms of wall-clock time compared to synchronous data parallel training and up to \textapprox27\% faster than comparable communication efficient algorithms while maintaining better task performance.
This speed-up is partly due to higher model FLOPs utilization, as we demonstrate.
By injecting delays into the communication between workers, we show that LayUp remains robust to stragglers while DDP and other methods degrade in performance.
Overall, LayUp provides a novel practical, straggler-robust alternative for distributed training without sacrificing accuracy.
\end{abstract}

\section{Introduction}

\begin{figure*}[htbp]
    \centering
    \includegraphics[width=\textwidth]{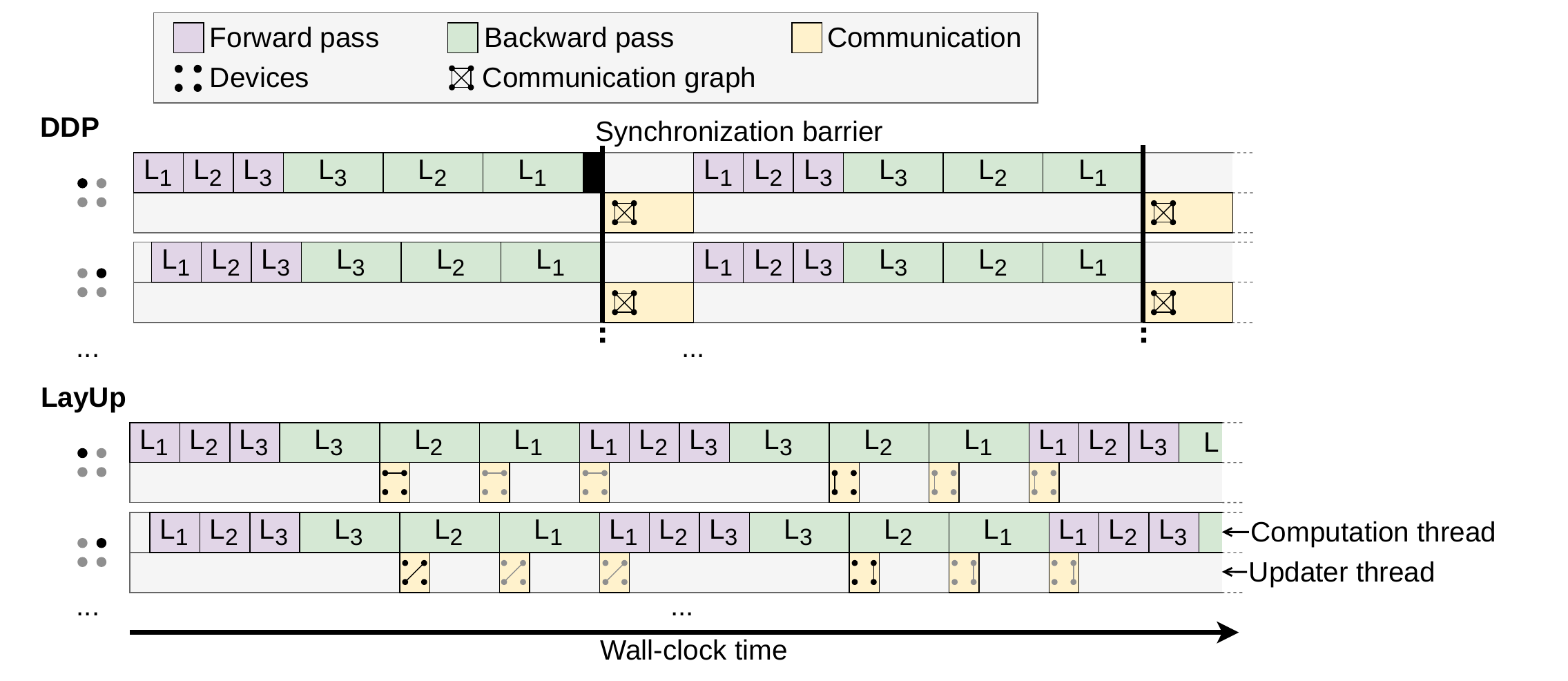}
    \caption{\textbf{Illustration of LayUp (bottom) and DDP (top).} 
        Each row shows the computation and communication threads separately, showing the layer-specific forward and backward passes and each time parameters are sent to other devices.
        The communication graph is shown each time a parameter update is communicated. 
        \textbf{DDP} waits for all the backward passes to complete using a synchronization barrier, and then does an all-reduce all-to-all communication to exchange parameter updates.
        \textbf{LayUp} allows devices to perform forward and backward passes in parallel, communicating updates from backward passes as soon as its available for each layer concurrently with the computation.
        For each device, the order of computations for a sample network with three layers denoted L1, L2 and L3, are shown. 
        Within each device, the computations for each layer are performed sequentially, whereas across threads, the dependencies are layer-wise.
    }
    \label{fig:async_partial_updates}
\end{figure*}


Modern deep learning models \citep{transformer, gpt3} require training at massive scale across many devices to achieve reasonable training times.
The most common approach is data-parallel mini-batch stochastic gradient descent following a distributed data parallel (DDP, \citet{DDP}) paradigm.
However, DDP incurs considerable communication and synchronization costs, leading to poor resource utilization and hindering horizontal scaling.

Several lines of work have explored possible solutions to this while trying to preserve model performance.
A natural alternative is asynchronous training~\cite{hogwild!}, where computation and communication overlap to reduce device idle time.
Asynchronous training is particularly helpful with heterogeneous devices to mitigate the overheads introduced by synchronization barriers. 
However, it often leads to worse task performance compared to synchronous training in practice~\cite{pmlr-v119-woodworth20a}.
Another popular approach is local stochastic gradient descent (Local SGD)~\cite{localSGD}, which reduces the frequency of synchronization between devices, thereby potentially speeding up the process. 
However, this can also negatively impact task performance or require additional memory when parameter drift (divergence) correction is applied.
A third strategy uses gossip-based communication to approximate the global average by letting only subsets of devices communicate.
Combining these strategies is theoretically possible, but can lead to high memory consumption or parameter drift.

In this work, we focus on the setting of asynchronous decentralized SGD.
Typical implementations use queues or buffers on each device to receive updates from peers, which increases memory usage, especially in the presence of stragglers -- workers lagging behind other workers due to computational or communication bottlenecks.
Moreover, updates are often applied only at the end of the backward pass or before the next iteration, even when earlier layer updates are already available.
Only applying the updates once the full backward pass has completed requires synchronization barriers, introducing a bottleneck and slowing down convergence.
Even if the updates are applied without a synchronization barrier (asynchronously ), waiting for the full backward pass to complete takes longer, and causes parameter drift in asynchronous methods.
Earlier work~\citep{agp, SGP} often ignores backward-pass timing and implicitly assume that all layers are updated at approximately the same time, which is not true for large models where output-layer updates are available long before input-layer updates.

To address these issues, we propose LayUp: asynchronous decentralized SGD with layer-wise updates, providing an alternative to data-parallel training.
LayUp applies layer-wise updates during backpropagation and allows concurrent `updater' threads (one per device) to directly update the parameters of a target device,
removing the need for queues or buffers to keep track of received updates.
Updater threads are responsible for the layer-wise lock-free updates of the parameters and communicate using random gossiping. 

Layer-wise updates reduce parameter drift between devices, but also introduce a bias.
However, we show both theoretically and empirically that the introduced bias does not adversely affect the model convergence or task performance.

Our method is orthogonal to model, tensor and pipeline parallelism \citep{zero_bubble}, and can be combined with any of these to enable training even larger models.


In summary the contributions of this paper are as follows:

\begin{enumerate}
    \item We introduce a new asynchronous distributed SGD-based training method called LayUp that allows overlapping communication and computation. LayUp makes incremental updates to the model parameters at a layer-wise granularity without using a locking mechanism; 
    \item We provide an upper bound on the gradient bias introduced by LayUp due to the layer-wise updates, using the elastic consistency framework;
    \item We provide theoretical convergence guarantees for the algorithm in the neighborhood of the optimal solution using the elastic consistency framework;
    \item We show that LayUp can achieve better task performance than similar methods in vision and natural language tasks while converging faster, being more robust to delays, and having higher hardware utilization than comparable DDP and communication-efficient algorithms. 
\end{enumerate}

\section{Related Work}

When training Deep Neural Networks (DNN) in distributed settings, communication quickly becomes the bottleneck and synchronization overheads lead to diminishing returns as the number of workers increases.
We briefly review asynchronous, local/periodic synchronization, and decentralized/gossip approaches, along with relevant convergence analyses.

Asynchronous algorithms have a long history, starting from \citet{baudet1978asynchronous} (See \citet{bertsekas2015parallel} for an overview).
One of the first papers to apply asynchronous updates to SGD was Hogwild! \citep{hogwild!}, which allowed multiple processes to perform SGD without any locking mechanism on shared memory.
Later, \citet{passm} proposed PASSM and PASSM+, where they partition the model parameters across the workers on the same device to perform SGD on the partitions, improving on Hogwild!
These methods focused on centralized shared-memory settings.

\citet{scaling_hogwild} extended Hogwild! and PASSM+ to a decentralized setting, allowing parameters or their partitions to be located on multiple devices and perform Local SGD \citep{localSGD} on them. 
To correct for the delayed gradients, \citet{delay_compensation} proposed a method with a gradient approximation at the current parameters.
Unlike these methods, we do not perform global averaging and do not need any gradient compensation scheme since layer-wise updates reduce parameter drift.

Synchronizing all workers after each iteration is costly. 
This can be mitigated by performing synchronization at a given frequency, which was first proposed in~\citet{localSGD} to our knowledge. 
Vanilla Local-SGD often under-performs in practice without careful tuning.
\citet{post-localSGD} proposed Post-Local SGD, a combination of Local-SGD followed by DDP, where finding a well balanced combination between them is not always obvious because of the different learning dynamics. 
To account for parameter drift among workers~\citet{slowmo} introduced SlowMo, which uses an outer optimization step at the cost of additional memory use. 
CO2~\cite{co2} overlaps parameters synchronization with outer optimization and introduces a penalty gap, needing 4x model size additional memory, to further reduce parameter drift across workers. 
DiLoCo~\cite{DiLoCo} uses AdamW for the inner optimization and SGD with Nesterov momentum for the outer optimization, but struggles to reach DDP performance.
\citet{EDiT} proposes a way to merge Local-SGD with the Zero~\citep{zero} and FSDP~\citep{fsdp} frameworks and introduce their own penalty term, making use of an additional buffer of the trained model size like SlowMo.
LayUp does not require extra buffers but relies on layer-wise updates to reduce parameter drift. 
Moreover, we do not introduce any additional hyper-parameters over DDP. 
We also show that our method achieves better task performance and faster convergence than other methods in many cases. 

A way to reduce communication volume is to use gossip-like communication, where a worker communicates only with a subset of peers at each iteration. 
The `consensus', defined as the implicit average of parameters across workers, is therefore approximated by peer-to-peer averaging. 
GoSGD~\cite{GoSGD} was one of the first random gossip-based~\cite{randomized_gossip} SGD method, proposing an asynchronous, push-sum~\cite{push_sum} based algorithm, scaling to ImageNet.
\citet{GoSGD-PA} alternated between global averaging and gossip averaging
, while~\citet{SGP} worked on directed graphs and implemented an average bias correction term, thus needing an extra buffer of the same size of the model being trained. 
To allow faster training, AD-PSGD~\cite{AD-SGD} allowed individual workers to operate completely asynchronously but enforced symmetric communication between at each iteration, hence doubling the  communication volume compared to GoSGD. 
LayUp builds on GoSGD, introducing partial layer-wise parameter updates that are lock-free and queueless, and applying them as soon as they become available during backpropagation.
Additionally, we theoretically bound the gradient bias, provide a proof of convergence for this setting, and show empirical results on a range of vision and language modeling tasks.



A standard framework for providing convergence in a setting such as ours is the Elastic Consistency framework \citep{nadiradze_2021_elastic}, which provides a theoretical framework for deriving convergence guarantees for a wide variety of distributed training methods.
In particular, \citep{nadiradze_2021_elastic} show that the elastic consistency framework can be used to analyze decentralized algorithms, which makes it applicable to LayUp.
\citet{mishchenko_2022_asynchronousa} proposed a method of ``virtual iterates'' to provide convergence guarantees independent of delays.
More recently, \citet{even_2024_asynchronous} proposed a unified framework for convergence analysis of distributed algorithms based on the AGRAF framework.
In our work, we formally characterize the gradient bias introduced by our method and establish an upper bound on its magnitude using the elastic consistency framework \citep{nadiradze_2021_elastic}.

\section{Methods}

\textbf{Optimization problem.} We consider $M$ workers collaborating to jointly minimize a function $f: \mathbb{R}^d \to \mathbb{R}$ over a data distribution $\mathcal{D}$
\begin{align}
    f(x) =  \frac{1}{M} \sum_{i=1}^M \EX_{S_k \sim  \mathcal{D}}\mathcal{L}(S_k,x^i) \,,
  \label{eq:avg_loss}
\end{align}

where $\mathcal{L}$ is the loss on sample $S_k$ computed on worker $i$ using local parameters $x^i$, and the $k$-th sample is exclusively used on device $i$ within a given epoch. 
We are looking for the solution to the  optimization problem 
$$x^* = \arg\min_x f(x),$$ 
where $x^*$ is the minimizer of the function $f$ over parameters $x$.

In the following, we describe how LayUp works -- see Figure~\ref{fig:async_partial_updates}. 
LayUp uses one thread per device performing the computation and another one `updater' thread that is dedicated to communicating and applying the updates.

At a given iteration, a forward pass is performed, say, on device $i$.
Once the forward pass is completed, we simultaneously start the corresponding backward pass on the same device.
Another forward pass might be running on device $j$ in parallel to device $i$.
The parameters of device $j$ will be updated by the updater thread located on device $i$ as soon as the updates for each layer are available. 
Device $j$ might therefore use those updates directly during its forward pass.
This allows devices to mix their parameters more frequently than if the parameters were only updated after all layers were processed.

At each iteration, a thread computes the loss $\mathcal{L}(S_k, \hat{x}^i)$, given the current mini-batch of data $S_k \in \mathcal{D}$ and the latest set of updated weights $\hat{x}^i$.
Since the algorithm works asynchronously, the weights $\hat{x}^i$ that are used could have been updated to a new version, $\Tilde{x}_i$, by any updater thread from other devices even while forward pass progresses, hence the updates: 
$$x^{i,l} \gets \Tilde{x}^{i,l} - \eta . \nabla \mathcal{L}(S_k, \hat{x}^{i,l})\,.$$
This can potentially lead to a forward pass on a model that doesn't correspond to any single complete backward pass.
We show both theoretically (Section~\ref{sec:convergence_analysis}) and empirically (Section~\ref{sec:results})   that the noise induced by the bias created through these interactions is well behaved, and beneficial to the algorithm performance in practice.


\begin{algorithm}[h!]
    \caption{ {\textsc{LayUp} 
    }}
    \label{alg:async}

        \vspace{0.7ex}
        {Training loop on device \textit{i}} 
        \vspace{0.7ex}
        \hrule
        \vspace{0.5ex}
        
        \begin{algorithmic}
            \STATE \textbf{Given:} \textit{$S_k \in \mathcal{D}$}: random sample , \textit{$\hat{x}^i$}: latest parameters view, \textit{$\eta$}: learning rate, weight: $w_i$ 
            \STATE \textbf{Forward pass}: \textit {Compute} $\mathcal{L}(S_k, \hat{x}^i)$  
            \FOR{ each layer $ l $} 
                    \STATE \textbf{Backward pass at layer $l$}: $\nabla \mathcal{L}(S_k, \hat{x}^{i,l})$ 
                    \STATE \textbf{Notify:} updater thread $i$;
            \ENDFOR
        \end{algorithmic}
        \vspace{1.5ex}
        \hrule
        \vspace{1.50ex}
        {Updater Thread \textit{i} \quad \textcolor{blue} {\sc{(Parallel to training loop)}}}
        \vspace{0.7ex}
        \hrule
        \vspace{0.7ex}
        \begin{algorithmic}
                \REPEAT
                    \STATE\textbf{Randomly select a peer:} \textit{j} $\sim$ \textit{Random (M-1)}
                    \STATE $w_i \gets \frac{w_i}{2}$
                    \FOR{ each layer $ l $}
                        \STATE \textbf{Wait for notification;}
                        \STATE \textbf{Local Update: }$x^{i,l} \gets {\Tilde{x}^{i,l}} - \eta . \nabla \mathcal{L}(S_k, \hat{x}^{i,l})$
                        \STATE \textbf{Communication:} send($x^{i,l}, j$) 
                        \STATE \textbf{Peer Update: }$x^{j,l} \gets \frac{w{j}}{w_i + w_{j}} \hat{x}^{j,l} + \frac{w_i}{w_i + w_{j}}x^{i,l}$  
                    \ENDFOR
                    \STATE $w_{j} \gets w_{j} + w_i$
                \UNTIL \textbf{{training ends}}
    \end{algorithmic}
\end{algorithm}

\subsection{Communication strategy}

To fully overlap computation and communication, an updater thread on each device is responsible for updating parameters. 
It does this by directly updating the parameters of a randomly selected peer.
%
Updater threads operate fully asynchronously and concurrently to the training loop, meaning they can potentially overwrite each other's updates. 
But this is unlikely to happen, as the probability of two devices selecting the same random peer goes exponentially fast to zero as the total number of devices $M$ grows. 
In case the same peer is randomly selected by two devices, the parameters sent from one of the peers will be skipped. 
But since each peer first updates its own parameters and then sends the update, no information is really lost.
Instead, the parameters from the peer whose earlier update was skipped will be sent again either by the same device or another device already having the information at the next turn.
Therefore, it only delays the propagation of information, rather than losing it.


To model how much a device contributes to solving the optimization problem in Equation~(\ref{eq:avg_loss}), it is assigned a weight $w_i = \frac{1}{M}$ at the beginning of training. 
The weight is updated according to the push-sum protocol~\cite{push_sum} when sending (halved) or receiving parameters (summed with the weight of the sending device). 
The selection of the receiving device is performed uniformly at random and ensures that each device contributes equally in expectation: $\EX[w_i] = \frac{1}{M}$.
In our experiments (Section~\ref{sec:results}) we used 
randomized selection of peers, where device $i$ communicates with a randomly chosen peer $j$ (as in randomized gossip).
Due to the lock-free nature of our algorithm, it is not guaranteed that all the weights for the push sum will be used, since some might be skipped due to contention (even though the parameter update itself is not lost, as described above).
But in practice, we observe that this does not affect the convergence of the training.
\subsection{Layer-wise updates}

Asynchronously updating parameters through overwriting violates several key assumptions of backpropagation leading to suboptimal convergence observed in different studies \citep{distributed_limit, delay_compensation}. 
This happens because losses and gradients are often calculated using inconsistent and outdated parameters.

To alleviate this problem, we perform layer-wise updates. 
That is, we update the layers as soon as the corresponding gradients for that layer are available from the backward pass. 
For example, in Figure~\ref{fig:async_partial_updates}, the second forward pass might have received partial parameter updates from the first backward pass as soon as they are available. 
Therefore, the parameters used during the second forward will differ from those used in the first because some layers of the model would already have been updated.
It is important to note that the updates happen without any locking mechanism and asynchronous to the backward pass.

\paragraph{Drift Reduction Analysis.}

Applying layer-wise updates in LayUp can reduce parameter drift, as we show here.
The drift reduction estimation is computed with respect to when updates are applied at the end of an iteration, similar to various previous asynchronous learning algorithms, e.g. \citet{hogwild!, scaling_hogwild, delay_compensation}.

To express this formally, we define the relative drift $D$ 
as the time delay between when the gradients become available and when they are actually used to update the model weights. 
The intuition for this is the following: the more the communication or application of updates is postponed, the more likely parameters across devices will drift apart.
The drift will only increase with time and accumulate across the layers.
Assuming that the time required to compute the gradients for each layer is identical and equal to $\frac{\beta T}{L}$, the relative drift is $D = \frac{\beta T (L+1)}{2}$, where $\beta T$ is the time required to perform a single backward pass.

To see this, we use the fact that the drift increases as we approach the output layer. 
At any layer $l$, the layer-wise drift/divergence is $D_l = \frac{\beta T}{L} l$.
Summing over the layers, we have
\begin{align}
    D &\;=\; \sum_{l=1}^{L} D_l \;=\; \frac{\beta T}{L} \sum_{l=1}^{L}l \;=\; \beta T\frac{(L+1)}{2} \nonumber \,.
\end{align}
Clearly, $D$ grows with the network's depth and the time required to perform one backward pass. 
Thus, the parameter drift between devices is expected to scale approximately linearly with the network depth, showing the advantage of partial layer-wise updates over complete block updates for large networks.

\section{Experimental setup}
\label{sec:results}

\begin{figure*}
    \centering
    \includegraphics[width=\textwidth]{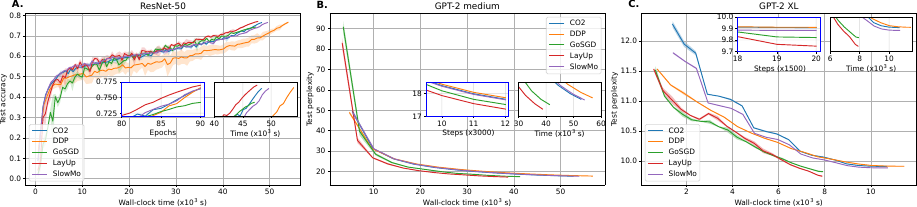}
    \caption{
    Learning curves for \textbf{A.} ResNet-50 test-accuracy on ImageNet-1k. \textbf{B.} GPT-2 Medium Pre-training on MiniPile. \textbf{C.} GPT-2 XL finetuning on Wikitext-103.
    Black insets: zoomed in view of the test perplexity in the last 55, 30, 6 ($\times 10^3)$ seconds in \textbf{A.}, \textbf{B.} and \textbf{C.} respectively.
    Blue insets: zoomed in view of how the test perplexity varies in the last 10 epochs, 2 ($\times 3000$) steps and 2 ($\times 1500$) steps rather than wall-clock time. 
    This is to demonstrate that while all the methods converge in approximately the same number of epochs/steps, the actual wall-clock time required for each epoch/step varies more significantly between methods.
}
    \label{fig:main-results}
\end{figure*}

\paragraph{Tasks.}
We evaluate LayUp on two vision tasks: ImageNet-1k~\citep{imagenet}, CIFAR-100~\citep{cifar100} 
and on sequence modeling tasks: large language Model (LLM) pre-training on Minipile~\citep{minipile} and finetuning on Wikitext-103~\citep{wikitext} datasets, both using the GPT architecture.

\paragraph{Setup.} We trained ResNet-50 on ImageNet-1k with a batch size of 256 per worker for 90 epochs and a linear decaying to zero learning rate after a warm-up phase of 2 epochs.
We trained both ResNet-18 and ResNet-50 on CIFAR-100 using a batch size of 128 per worker and a cosine learning rate scheduler.\\
For sequence modeling, we pre-trained GPT-2 medium (400M) on Minipile and finetuned GPT-2 XL (1.6B) on Wikitext103 using respectively a batch size of 48 and 4 per worker and cosine learning rate scheduler. 
We also tested LayUp on a sentiment analysis task with recurrent neural network architecture. 
Results for this can be found in Table~\ref{tab:results-imdb-multiseed} in Appendix~\ref{appendix:Imdb}.
Details of the different hyper-parameters are to be found in Appendix~\ref{appendix:hyperparams}. 

\paragraph{Hardware.} The experiments were performed on 3 hardware configurations.
    \textbf{C1}: 3 NVIDIA A100-PCIe 80GB GPUs with two AMD EPYC CPUs sockets of 64 cores each; 
    \textbf{C2}: 8 NVIDIA A100-SXM4 40GB GPUs with two AMD EPYC CPUs sockets of 24 cores each;
    \textbf{C3}: 4 NVIDIA H100-SXM5 94GB GPUs with two AMD EPYC CPUs sockets of 32 cores each.
\label{ini_machine_specs}

We used configuration C1 for all vision tasks, C2 for pre-training and C3 for finetuning. 
This combination of different hardware specifications, model, and dataset sizes will help to showcase the performance of the algorithms in different scenarios.

\textbf{Metrics.} 
For vision tasks, we record the achieved accuracy for the tasks, the wall-clock time to reach convergence (TTC), and the wall-clock time to reach a target accuracy (TTA). 
We record the achieved perplexity and training time for sequence modeling and finetuning tasks.

\textbf{Baseline.} The performance on these tasks is primarily compared to de-facto training algorithm DDP but also to CO2~\citep{co2} and SlowMo~\citep{slowmo} representing recent alternatives to DDP that relax synchronization.
For each task, we carefully performed extensive hyper-parameters search to obtain the best performance for each algorithm.

LayUp was implemented with the C++ front-end of the Torch library (LibTorch)~\citep{pytorch} to allow real multi-threading\footnote{When implementing our code, Python had not released a version allowing true multi-threading yet.}. 
SlowMo and CO2 implementations were also re-implemented using C++/LibTorch, using the available Python implementation as a reference.
The published CO2 implementation does not include the penalty gap correction, our implementation follows the same approach\footnote{The penalty gap correction is expected to affect only final task-performance, and not the convergence speed.}. 
Further implementation details are found in Appendix~\ref{appendix-implementation}. We submit the code as supplementary material as well.

\label{results}

\section{Results}
\subsection{Vision tasks}
\label{sec:vision}
We first test LayUp on small-scale vision tasks to validate stability and convergence.
Our results show that LayUp  successfully converges (Figure~\ref{fig:main-results}) in all cases to good solutions.
Table~\ref{tab:results-vision-best_acc}
shows that LayUp also achieves the highest accuracy on both CIFAR-100 and ImageNet-1K among the algorithms we tested.
In both cases, LayUp is faster than synchronous DDP: \textapprox16\% and \textapprox12\% on CIFAR-100/ResNet-50 and Imagenet-1k respectively. 
Since CO2 and SlowMo converge prematurely to a suboptimal solution that is almost 1-2\% points lower than LayUp in terms of accuracy, they converge at about the same time (ImageNet-1K) or earlier (CIFAR-100) than LayUp.

Instead, if we measure the time to a fixed target accuracy chosen to be the best accuracy achieved by the worst performing algorithm (see Table~\ref{tab:results-vision-tta}), then LayUp reaches the target accuracy \textapprox 7\% faster for CIFAR-100/ResNet-50 and \textapprox 4\% faster for ImageNet-1k/ResNet-50 compared to the best alternative distributed algorithm.
See Tables~\ref{tab:results-cifar100-tta-resnet18} and~\ref{tab:results-cifar100-best_acc-resnet18} in Appendix~\ref{appendix:vision-results} for ResNet-18 results. 

\begin{table}[ht]
    \centering
    \caption{Comparison of the algorithms on vision tasks 
    based on ResNet-50 convergence accuracy, TTC, and epoch at which the accuracy is achieved, each averaged over 3 runs. 
*For ImageNet-1k they all reach their peak accuracy at the 90th epoch.
}
    \resizebox{\columnwidth}{!}{%
    \begin{tabular}{@{}llr@{~$\pm$~}l r@{~$\pm$~}l r@{~$\pm$~}l@{}}
\toprule
        Task &  Method & \multicolumn{2}{c}{Convergence } & \multicolumn{2}{c}{TTC (seconds) $\downarrow$} & \multicolumn{2}{c}{Epochs $\downarrow$} \\ 
         &   & \multicolumn{2}{c}{ accuracy $\uparrow$} & \multicolumn{2}{c}{} & \multicolumn{2}{c}{} \\ 
          &          & mean  & std                & mean    & std              & mean & std \\
\midrule
CIFAR-100 
          & DDP & 78.27 & {\pmsize 0.16} & 923.00 & {\pmsize 47.09}  & 103  & {\pmsize 5} \\
          & CO2 & 77.94 & {\pmsize 0.22} & 847.67   & {\pmsize 47.52} & 102  & {\pmsize 5} \\
          & SlowMo   & 77.82 & {\pmsize 0.20} & \textbf{843.00} & \textbf{\pmsize 31.32}  & \textbf{97}  & \textbf{\pmsize 4} \\
          & GoSGD   & 76.47 & {\pmsize 0.39} & 1059.33  & {\pmsize 19.14} & 105  & {\pmsize 2} \\ 
          & AD-PSGD & 77.75 & 0.21 & 708.33 & 19.30 & 91 & 2\\
          & LayUp (ours) & \textbf{78.46} & \textbf{\pmsize 0.18} & 844.67 & \pmsize 17.62  & 106  & \pmsize 2 \\
\midrule
    %
    %
ImageNet-1k 
          & DDP & 76.57 & {\pmsize 0.30} & {\pmsize 53930.67} & 1609.37 & 90 & 0* \\
          & CO2   & 76.42 & {\pmsize 0.06} & 48092.33   & {\pmsize 847.09} &  90 & 0* \ \\
          & SlowMo   & 76.43 & {\pmsize 0.08} &  49169.33  & {\pmsize 833.46} &  90 & 0* \ \\
          & GoSGD & 74.19 & {\pmsize 0.49} & \textbf{47005.67}   & \textbf{\pmsize 451.90} &  90 & 0* \ \\
          & AD-PSGD & 76.04 & 0.29 & 50995.33 & 301.60 & 90 & 0*\\
          & LayUp (ours) & \textbf{76.97} & {\pmsize \textbf{0.17}} & {47463.00}   & {\pmsize 109.42} &  90 & 0* \ \\
\bottomrule
    \end{tabular}
    }
    \label{tab:results-vision-best_acc}
\end{table}

\begin{table}[ht]
    \centering
    \caption{Comparison of the algorithms 
    based on ResNet-50 time to reach accuracy (TTA): \textbf{76.07\%} for CIFAR100,  \textbf{73.56\%} for ImageNet-1k and the number of epochs to reach the target, each averaged over 3 runs.}
    \resizebox{\columnwidth}{!}{%
    \begin{tabular}{@{}llr@{~$\pm$~}l r@{~$\pm$~}l r@{~$\pm$~}l@{}}
\toprule
        Task &  Method  & \multicolumn{2}{c}{TTA (seconds) $\downarrow$} & \multicolumn{2}{c}{Epochs $\downarrow$} \\ 
          &          & mean    & std              & mean & std \\
\midrule
CIFAR-100 
          & DDP &  803.57 & {\pmsize 3.85}  & 89  & {\pmsize 1} \\
          & CO2 &  681.33   & {\pmsize 2.08} & 83  & {\pmsize 0} \\
          & SlowMo   &  760.33 & {\pmsize 11.02}  & 88  & {\pmsize 1} \\
          & GoSGD   &  694.67 & {\pmsize 49.69} & 89  & {\pmsize 6} \\ 
          &  AD-PSGD & 651.31 & 2.77 & 83 & 0\\
          & LayUp (ours) & \textbf{634.67} & \textbf{\pmsize 3.21}  & \textbf{80}  & \textbf{\pmsize 0} \\
\midrule
ImageNet-1k 
          & DDP & {\pmsize 51949.67} & 2243.13 & 87 & {\pmsize 1} \\
    
          & CO2   &  45950.33   & {\pmsize 565.40} & 86 & {\pmsize 1} \\
          & SlowMo   &   47154.33  & {\pmsize 1129.40}& 86 & {\pmsize 1}  \\
          & GoSGD &  45967.00   & \pmsize 918.50 & 88 & {\pmsize 1}\\
          & AD-PSGD & 48905.67 & 479.74 & 86 & 1\\
          & LayUp (ours)  & \textbf{43950.67}   & \textbf{\pmsize 346.43} & \textbf{83} & \textbf{\pmsize 1}\\
\bottomrule
    \end{tabular}
    }
    \label{tab:results-vision-tta}
\end{table}

\subsection{Sequence modeling task}
\label{sec:sequence}

We test LayUp on larger scale language modeling tasks, pre-training a GPT-2 Medium (400M) model and fine-tuning a GPT-2 XL (1.6B) model.
Table~\ref{tab:results-pretraining-pxb} presents the results for both pre-training and finetuning tasks. 
LayUp reaches the lowest perplexity in both cases.
It converges \textapprox5\% faster than DDP in pre-training and \textapprox32\% in finetuning and \textapprox27\% faster than SlowMo on pre-training, but lags slightly ($<0.5\%$) behind GoSGD in finetuning convergence speed.

The learning curve in Figure~\ref{fig:main-results}B shows that the perplexity of both decentralized algorithms is high for the first few thousand training steps compared to centralized algorithms. 
It is likely the period where a consensus among workers is yet to be found.
But they quickly reach and maintain a more stable training trajectory.
This trend is however not observable in fine-tuning (Figure~\ref{fig:main-results}C). 
This is likely because the parameters of the pre-trained model are not far from reaching a consensus.

\begin{table}[ht]
    \centering
    \caption{Algorithms performance on sequence modeling over 3 runs: pretraining of GPT-2 Medium on Minipile and Finetuning of GPT-2 XL on Wikitext-103.
    }

    \resizebox{\columnwidth}{!}{%
    \begin{tabular}{@{}llr@{~$\pm$~}l r@{~$\pm$~}l r@{~$\pm$~}l@{}}
\toprule
        Architecture & Method & \multicolumn{2}{c}{Perplexity $\downarrow$} & \multicolumn{2}{c}{Time (in seconds) $\downarrow$}\\ 
          &          & mean  & std\\
\midrule 
GPT-2 Medium 
          & DDP &17.85 & {\pmsize 0.01} &
          {\pmsize 40658.05} & 324.68  \\
    
          & CO2   & 17.75 & {\pmsize 0.01} & 51508.00   & {\pmsize 761.86} \\
          & SlowMo   & 17.72 & {\pmsize 0.02} &  52687.00  & {\pmsize 72.75} \\
          & GoSGD & 17.52 & {\pmsize 0.01} & 41147.67   & \pmsize 557.10 \\
          &  AD-PSGD & 26.69 & 0.04 & 38093.67& 8.96 \\
          & LayUp (ours) & \textbf{17.32} & \textbf{\pmsize 0.01} & \textbf{38624.33}   & \pmsize \textbf{222.46} \\
          \midrule
          GPT-2 XL 
          & DDP & 9.91 & {\pmsize 0.01} &
          {\pmsize 11283.33} & 0.01 \\
    
          & CO2   & 9.91 & {\pmsize 0.01} & 9789.33 & {\pmsize 47.38  } \\
          & SlowMo   & 9.88 & {\pmsize 0.00} &  9791.33  & {\pmsize 7.51} \\
          & GoSGD & 9.82 & {\pmsize 0.01} & \textbf{7652.33}  & \textbf{\pmsize 25.32} \\
          &  AD-PSGD & 10.52 & 0.00 & 7860.00 & 11.53\\
          & LayUp (ours) & \textbf{9.75} & \textbf{\pmsize 0.02} & {7688.33}   & {\pmsize 38.89} \\
          
\bottomrule
    \end{tabular}
    }
    \label{tab:results-pretraining-pxb}
\end{table}



\subsection{Model Flops Utilization}

To understand whether overlap of computation and communication with incremental updates helps LayUp utilize the GPUs more effectively, we measured the model FLOPs utilization (MFU) when training with LayUp compared to other algorithms.
The MFU measures the ratio of achieved FLOPs to the theoretical peak FLOPs of the underlying hardware, as introduced by~~\citet{palm}.
All models were trained with 32-bit floating point precision.

Among the algorithms, LayUp achieves the highest hardware utilization for the pre-training task, but lags slightly ($<0.3\%$) behind GoSGD for the finetuning task (see Table~\ref{tab:mfu_pretrain_pxb}). 
This confirms that allowing threads to operate independently without synchronization maximizes resource usage, and reduces waiting times due to synchronization (see Figure~\ref{fig:async_partial_updates}).

\begin{table}[ht]
    \centering
    \tiny
    \caption{Comparison of Model FLOPs Utilization (MFU) on Sequence modeling tasks.}

    \resizebox{\columnwidth}{!}{%
    \begin{tabular}{@{}llr@{~$\pm$~}l r@{~$\pm$~}l}
    \toprule
     Architecture & Method & \multicolumn{2}{c}{MFU (in \%) $\uparrow$} \\ 
                     &                 & mean & std \\ \midrule
                  GPT-2 Medium (pre-training)  & DDP        & 68.86  & 0.55  \\[0.5ex]
                  & CO2          & 53.17  & 0.78   \\
                  &SlowMo          & 51.95  & 0.08  \\
                  & GoSGD       & 67.94 &  0.93  \\
          & AD-PSGD & 73.49 & 0.02 \\
                  & LayUp (ours)       & \textbf{72.36} & \textbf{0.41}  \\
                          \midrule  
    GPT-2 XL (finetuning)  & DDP     & 42.29 & 0.32  \\
                   &CO2          & 49.38 & 0.24  \\
                    &SlowMo          & 49.37 &  0.04  \\
                    & GoSGD      & \textbf{63.17} & \textbf{0.21}  \\
          & AD-PSGD & 61.50 & 0.09\\
                    & LayUp (ours)       & {62.87} & {0.32}  \\
                          \bottomrule
    \end{tabular}
    }
    \label{tab:mfu_pretrain_pxb}
\end{table}


\subsection{Robustness to delays}
\label{delay}
In Sections~\ref{sec:vision} and \ref{sec:sequence} we studied the case where all processes run across multiple GPUs on the same node.
However, a more interesting setting is one where devices don't operate at the same speed, which can for example be the case in heterogeneous clusters. 
A related setting is when there are communication delays between devices due to bandwidth saturation.
To emulate both these scenarios using a controlled setting, we ran experiments where one device (straggler) is caused to operate slower by artificially adding delays by forcing it to idle.
The idle time is chosen to be a multiple of the time required by the algorithm to perform one forward and one backward pass (see Table~\ref{tab:BP-fwd-bwd-timing-compare}).
The delays are expressed in terms of the number of iterations the straggler lags behind the remaining devices.

\begin{figure*}[ht]
    \centering
      \includegraphics[width=\linewidth]{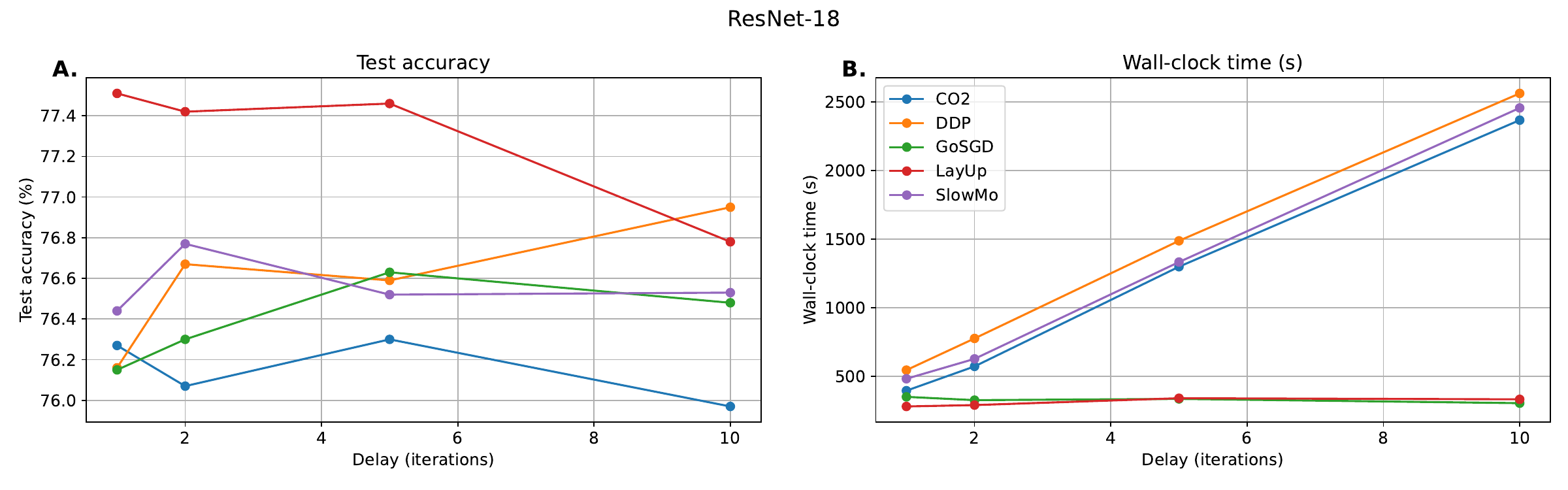}
      \caption{Effect of stragglers on training on CIFAR100 / ResNet-18.  \textbf{A.} test accuracy as a function of delay. All the training methods remain relatively unaffected by delay. \textbf{B.} training time as a function of delay. Only GoSGD and LayUp are robust to stragglers, while all the other methods deteriorate significantly.}
      \label{fig:delay_time}
\end{figure*}

We found that the test accuracy, and especially the training time of LayUp is mostly unaffected by stragglers as shown in Figure~\ref{fig:delay_time}. 
On the other hand, the training time for all other methods increases significantly in the presence of delays.  
This is a direct consequence of the asynchronous nature of LayUp, and the parameter-drift mitigating layer-wise updates.
Faster devices can continue the model execution without having to wait for the slowest device. 
The straggler keeps receiving updates from faster devices hence reducing parameter drift.

\section{Theoretical analysis of convergence}
\label{sec:convergence_analysis}
To analyze the convergence of LayUp, we use the elastic consistency framework established by~\citet{nadiradze_2021_elastic}, which allows to derive convergence properties of asynchronous decentralized algorithms.
But unlike elastic consistency, we analyze the convergence in the presence of biased gradients which are introduced due to the layer-wise updates.


We consider a distributed system with $M$ workers capable of communicating with each other through message (parameters) passing. 

\subsection{Convergence results}
We provide convergence guarantees for LayUp for both non-convex and convex cases. 
The detailed proofs are provided in Appendix \ref{appendix:proofs}.

To establish the convergence of LayUp, it is important to analyze the contribution of  layer-wise updates to the algorithm. 
By allowing updates to be fully asynchronous and layer-wise, there is possibility of a device to perform a backward pass on different weights than those on which the corresponding forward pass was performed. 
The resulting gradients can therefore be biased, defined as the difference between true gradients and gradients used during the backward pass.
We start by deriving an upper bound on the bias using the elastic bound assumption.

\begin{lemma}
\label{bias_bound}
Under the assumption  of smoothness and elastic consistency bound (Assumptions \ref{smoothness} and \ref{elastic_consistencey} in appendix~\ref{appendix:assumption}), the second moment of the gradients bias induced by LayUp (Algorithm Listing \ref{alg:async}) is bounded.
\begin{align}
    \forall x \in \mathbb{R}^d, \EX \norm{b(x)}^2 \leq 4K_b^2\eta^2B^2 \,,
\end{align}
\end{lemma}

where $K_b > 0$ is a Lipschitz constant, $B>0$ the elasticity constant (see Appendix~\ref{appendix:assumption}).

Given a bound on the gradient bias, we can show, following ~\citet{nadiradze_2021_elastic}, that the expected average of the norm of the gradients at the consensus shrinks over time, effectively reaching a minimum.  
\begin{theorem}
    \label{non_convex_theorem}
    Considering the SGD iterations of algorithm \ref{alg:async} and satisfying assumptions biased stochastic gradients, bounded variance, and elastic consistency bound (Assumptions \ref{bias_equation}, \ref{variance_bound}, \ref{elastic_consistencey} in Appendix~\ref{appendix:assumption}). Given a smooth, non-convex objective function $f$, with unknown minimum at $x^*$ and learning rate $\eta = \frac{\sqrt{M}}{\sqrt{T}}$, where $T \ge 144K_f^2M$ is the number of iterations.
    \begin{align*}
\frac{1}{T} \sum_{t=0}^{T-1} \EX \norm{\nabla f (\bar{x}_t)}^2 = \bigO\left( \frac{4 (f(\bar{x}_0) - f(x^*))}{\sqrt{TM}}  +\frac{4K\sigma^2}{\sqrt{TM}} \right. \\+
\left. \frac{60 K^3 B^{'2}\tau_{max}^2 M\sqrt{M}}{T\sqrt{T}} + \frac{20K^2B^{'2}\tau_{max}^2M}{T} \right) \notag \,,
    \end{align*}
\end{theorem}

\label{sec:convergence-theory}
where $K_f$ is a Lipschitz constant and $K = \max\{K_f, K_b\}$,  $\tau_{max}$ the maximum message delay of the system and $\sigma > 0$ is the noise arising from performing SGD steps (see Appendix~\ref{appendix:assumption}). The elastic bound $B = B^{'}\tau_{max}$.

This indicates that LayUp converges at a rate similar to DDP in both non-convex, $\bigO\left(\frac{1}{\sqrt{TM}}\right)$, and strongly-convex cases, $\bigO\left(\frac{1}{TM}\right)$. 
It shows that the bias introduced by the layer-wise updates is indeed well-behaved and does not affect convergence.


\section{Discussion}


In this work, we introduced LayUp, an asynchronous decentralized SGD method for training deep neural networks.
LayUp uses concurrent updater threads to apply layer-wise updates during backpropagation as soon as they are available, reducing parameter drift without requiring extra buffers.
LayUp addresses key limitations of standard synchronous SGD including data-parallel training by overlapping communication and computation while the layer-wise updates make it robust to stragglers.

The experimental results on vision and language modeling tasks demonstrate that LayUp achieves better task performance in all cases we tested, while converging up to \textapprox32\% faster than DDP and up to \textapprox27\% faster than comparable decentralized methods.
We demonstrate that LayUp leads to higher model flops utilization, which contributes to the observed speedups.
We also show that LayUp is robust to injected delays, remaining stable when stragglers are present while DDP and other baselines degrade in performance.

Our theoretical analysis, using the elastic consistency framework, provides an upper bound on the gradient bias introduced by layer-wise updates and also guarantees convergence. 
This offers a foundation for understanding the behavior of our asynchronous decentralized SGD approach.

Overall, this work presents a promising direction for distributed training of deep neural networks.
LayUp can enable more efficient training of large-scale models, particularly in heterogeneous environments. 
Moreover, LayUp has the potential to provide a more scalable, efficient and robust method for training large models.
Further research could explore extensions to larger models, additional tasks, and more diverse hardware setups to fully realize the potential of this training paradigm.



\section*{Impact Statement}

This paper presents work whose goal is to advance the field of Machine Learning. There are many potential societal consequences of our work, none which we feel must be specifically highlighted here.

\section*{Acknowledgements}
Lukas König and David Kappel were funded by the  German Federal Ministry of Research, Technology and Space (BMFTR) projects ESCADE (01MN23004D) and SAIL (grant no. NW21-059A) respectively.
The authors gratefully acknowledge the Gauss Centre for Supercomputing e.V. (www.gauss-centre.eu) for funding this project by providing computing time on the GCS Supercomputer JUWELS at Jülich Supercomputing Centre (JSC). The authors gratefully acknowledge the computing time made available to them on the high-performance computer at the NHR Center of TU Dresden. This center is jointly supported by the Federal Ministry of Research, Technology and Space of Germany and the state governments participating in the NHR (www.nhr-verein.de/unsere-partner).



\bibliographystyle{icml2026}
\bibliography{references}
\newpage

\appendix
\onecolumn
\setcounter{figure}{0}
\setcounter{table}{0}

\renewcommand{\thefigure}{A\arabic{figure}}
\renewcommand{\thetable}{A\arabic{table}}

\section{Additional results}
\label{appendix:results}

\subsection{Vision Tasks}
\label{appendix:vision-results}
We now analyze the time it takes the different algorithms to reach a given target accuracy, time to target (TTA). The models are trained till convergence and the target accuracy is chosen to be the accuracy reached by the worst performing algorithm. \\
We observe reaches the the target accuracy the fastest for both Imagenet-1k and CIFAR-100 on ResNet-50 up to $\sim$ 11\% faster than DDP. LayUp also needs the least number of epochs meaning it is more sample efficient.

\begin{table}[ht]
    \centering
    \caption{Comparison of the algorithms 
    based ResNet-18 on time to reach accuracy (TTA): \textbf{75.56\%}, 
     and the number of epochs to reach the target for 3 runs on CIFAR100.}
    \begin{tabular}{@{}llr@{~$\pm$~}l r@{~$\pm$~}l r@{~$\pm$~}l@{}}
\toprule
        Architecture &  Method  & \multicolumn{2}{c}{TTA (seconds) $\downarrow$} & \multicolumn{2}{c}{Epochs $\downarrow$} \\ 
          &          & mean    & std              & mean & std \\
\midrule
ResNet-18
          & DDP &  265.63   & {\pmsize 1.45} & 90  & \textbf{\pmsize 1} \\
          & CO2 & \textbf{203.33}  & \textbf{\pmsize 1.15}  & 86  & {\pmsize 1} \\
          & SlowMo   &  227.33  & {\pmsize 1.15}  & 90  & {\pmsize 1} \\
          & GoSGD   & 233.67 & {\pmsize 9.45}  & 93  & {\pmsize 5} \\
          & LayUp(ours) & 206.67  & \pmsize 5.69 & \textbf{79}  & \textbf{\pmsize 2} \\
\bottomrule
    \end{tabular}
    \label{tab:results-cifar100-tta-resnet18}
\end{table}

\begin{table}[ht]
    \centering
    \caption{Comparison of the algorithms on CIFAR100 
    based on ResNet-18 convergence accuracy, TTC, and epoch at which the accuracy is achieved on CIFAR-100 over 3 runs}
    \begin{tabular}{@{}llr@{~$\pm$~}l r@{~$\pm$~}l r@{~$\pm$~}l@{}}
\toprule
        Architecture &  Method & \multicolumn{2}{c}{Convergence} & \multicolumn{2}{c}{TTC (seconds) $\downarrow$} & \multicolumn{2}{c}{Epochs $\downarrow$} \\ 
         &   & \multicolumn{2}{c}{accuracy $\uparrow$} & \multicolumn{2}{c}{} & \multicolumn{2}{c}{} \\ 
          &          & mean  & std                & mean    & std              & mean & std \\
\midrule
ResNet-18
          & DDP & 76.57 & {\pmsize 0.02} & 300.67   & {\pmsize 2.89} & 103  & \pmsize 1 \\
          & CO2 & 76.64 & {\pmsize 0.26} & \textbf{238.00}  & \textbf{\pmsize 9.17}  & \textbf{100}  & \textbf{\pmsize 5} \\
          & SlowMo   & 76.57 & {\pmsize 0.04} & 256.67  & {\pmsize 1.15}  & 101  & {\pmsize 1} \\
          & GoSGD   & 75.86 & {\pmsize 0.40} & 258.00 & {\pmsize 17.58}  & 102  & {\pmsize 6} \\
          & LayUp(ours) & \textbf{77.74} & \textbf{\pmsize 0.25} & 267.33   &\pmsize 7.02 & 102  & \pmsize 3 \\
\bottomrule
    \end{tabular}
    \label{tab:results-cifar100-best_acc-resnet18}
\end{table}

\subsection{Sentiment Analysis task}
\label{appendix:Imdb}
For demonstrating LayUp training on sequence modeling, we evaluated an LSTM network on the IMDb sentiment analysis dataset~\citep{imdb_dataset}.
Sentiment analysis is the task of classifying the polarity of a given text \citep{sentiment_analysis}. 
 We used a 2-Layer LSTM network with 256 hidden dimensions to evaluate this task.
We trained the network until convergence using the Adam optimizer with an initial learning rate of $1\cross10^{-3}$.
Results are shown in Table~\ref{tab:results-imdb-multiseed}.

We observe that both LayUp and DDP perform similarly because of the small number of epochs to needed to reach reach convergence. 

\begin{table}[ht]
    \centering
    \caption{Comparison of LayUp, DDP based on convergence accuracy, time to reach convergence (TTC), and epoch at the accuracy achieved for 3 runs on IMDb}
    
    \begin{tabular}{@{}llr@{~$\pm$~}l r@{~$\pm$~}l r@{~$\pm$~}l@{}}
\toprule
Architecture  &  Method   & \multicolumn{2}{c}{Convergence} & \multicolumn{2}{c}{TTBA (seconds) $\downarrow$} & \multicolumn{2}{c}{Epochs $\downarrow$} \\  
  &     & \multicolumn{2}{c}{accuracy $\uparrow$} & \multicolumn{2}{c}{} & \multicolumn{2}{c}{} \\  
             &          & mean & std & mean & std & mean & std \\ \midrule
LSTM 	     
            & DDP      & 84.60  & {\pmsize 0.51}   & 54.0  & {\pmsize 2}  & 6  & \pmsize 1 \\
             & LayUp (ours) & \textbf{85.15}  & {\pmsize 0.15}   & \textbf{49.3}  & \textbf{\pmsize 9}  & 6  & \pmsize 1 \\
             
\bottomrule
\end{tabular}
    \label{tab:results-imdb-multiseed}
\end{table}

\subsection{Models disagreement measurements}

\begin{figure}
    \centering
    \includegraphics[width=0.8\textwidth]{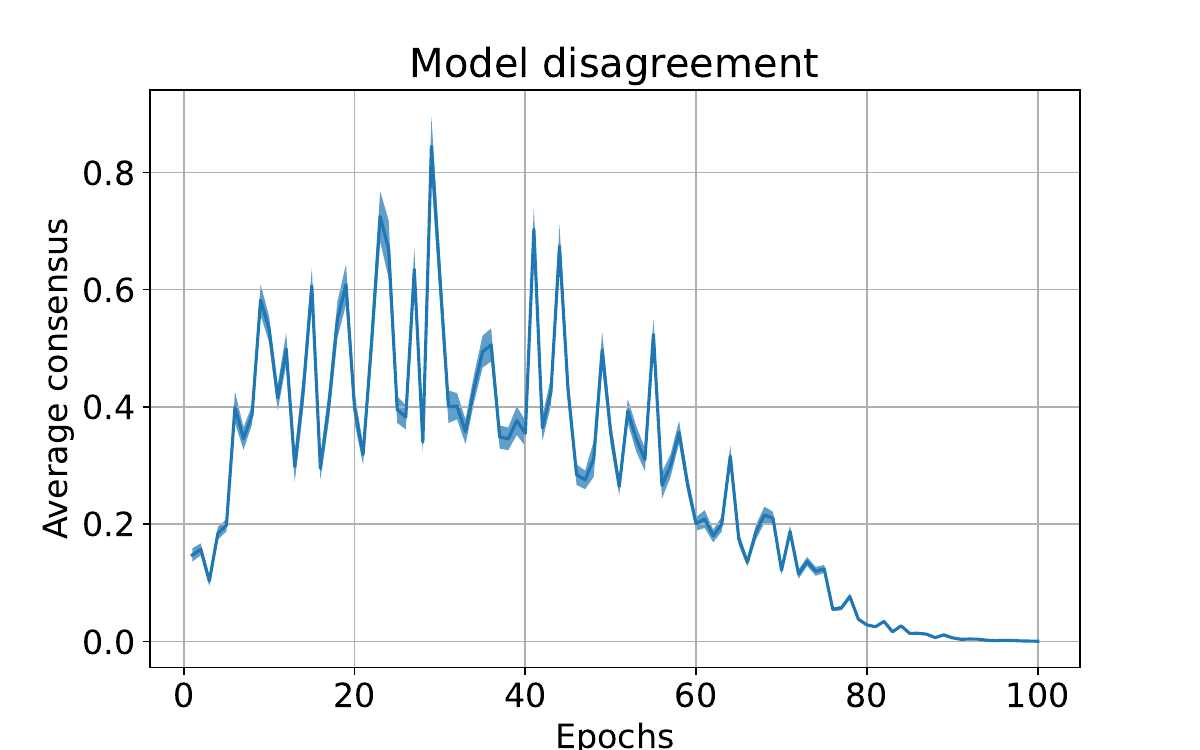}
    \caption{Model disagreement between workers when training resnet18 on cifar100.}
    \label{fig:consensus}
\end{figure}

We empirically track the model disagreement between the different throughout training. We can see in figure~\ref{fig:consensus} that the model disagreement is bounded, in this case $\leq 1$, and goes to zero as the training ends ($T_{max} = 100 $). This validates the theory. 

\subsection{Time measurements}
\label{apx:speed}

Here we provide the results of a small-scale experiment on the
timing measurement of forward and backward passes for CIFAR-100 with batch size 128 in table~\ref{tab:BP-fwd-bwd-timing-compare}. As expected, a single backward pass requires $\sim 2\times$ than that of a single forward pass. Extensive experiments on this is provided by \citet{kumar2021doing}

\begin{table}[ht!]
    \centering
    \caption{Timing measurement of forward and backward passes for CIFAR-100 with batch size 128. Averaged over all batches for 15 epochs.}
    \begin{tabular}{@{}lr@{~$\pm$~}l r@{~$\pm$~}l@{}}
    \toprule
    Network architecture & \multicolumn{2}{c}{Forward pass (s)} & \multicolumn{2}{c}{Backward pass (s)} \\ 
          & mean  & std                & mean    & std \\
\midrule
ResNet-18 & 0.0049 & {\pmsize 1e-04} & 0.0102  & {\pmsize 1e-04} \\
ResNet-50 & 0.0166 & {\pmsize 5e-05} & 0.0299  & {\pmsize 4e-05} \\
\bottomrule
    \end{tabular}
    \label{tab:BP-fwd-bwd-timing-compare}
\end{table}

\subsection{Hyperparameters for the experiments}
\label{appendix:hyperparams}
Now we present hyperparameters used for the various tasks. To this end, we used the Bayes' approach~\cite{bayes_hyperparameter} implemented by wandb~\cite{wandb}. 
For all CO2 experiments, following the original paper, the momentum clipping threshold is kept as 1.0 for all experiments.

\begin{table}[h]
\centering
\caption{Experiments hyperparameters for Imagenet-1k training}
\label{tab:hyperparams-imagenet-appendix}
\begin{tabular}{@{}llllll@{}}
\toprule
                     & DDP           & CO2           & SlowMo        & LayUp / GoSGD  \\ \midrule
batch\_size          & 256            & 256            & 256            & 256      \\
optimizer            & SGD         & SGD         & SGD         & SGD  \\
lr                   & 0.3       & 0.3        & 0.3        & 0.1\\
Cosine scheduler (T\_max)               & 90         & 90         & 90         & 90 \\
warm\_up\_epochs & 2             & 2             & 2             & 2  \\
warm\_up lr          & 0.1             & 0.1             & 0.1       	     & 0.03 \\
weight\_decay        & 0.0001             & 0.0001             & 0.0001             & 0.0001      \\
out\_freq            & -             & 48            & 48            & -      \\
tau                  & -             & 12             & 12             & -      \\
\bottomrule
\end{tabular}
\end{table}

\begin{table}[h]
\centering
\tiny
\caption{Experiments hyperparameters for CIFAR-100}
\label{tab:hyperparams-cifar100-appendix}
\begin{tabular}{@{}lllllllllll@{}}
\toprule
                     & \multicolumn{2}{c}{DDP}           & \multicolumn{2}{c}{CO2 }          & \multicolumn{2}{c}{SlowMo}        & \multicolumn{2}{c}{LayUp}   & \multicolumn{2}{c}{GoSGD}\\ \midrule
                     &ResNet-18 & ResNet-50&ResNet-18 & ResNet-50&ResNet-18 & ResNet-50&ResNet-18 & ResNet-50&ResNet-18 & ResNet-50\\
batch\_size          & 128   & 128    & 128   & 128    &   128   & 128 &128  & 128  & 128 & 128\\
optimizer            & SGD         & SGD         & SGD         & SGD  & SGD & SGD & SGD  & SGD & SGD & SGD\\
lr                   & 0.045       & 0.045        & 0.045        & 0.045 & 0.045       & 0.045 & 0.035      & 0.035 & 0.045 & 0.022\\
Cosine scheduler (T\_max)               & 100         & 100         & 100         & 100 &  100   & 100& 100  & 100 & 100 & 100\\
warm\_up\_epochs & 0             & 0             & 0             & 0  & 0             & 0 & 5             & 5  & 5 & 5\\
warm\_up lr          & 0             & 0             & 0       	     & 0 & 0             & 0 & 0.012       & 0.01 & 0.015 & 0.0073\\
weight\_decay        & 0.005            & 0.005             & 0.005             & 0.005 & 0.005   & 0& 0.003  & 0.005& 0.005 & 0.014 \\
out\_freq            & -             & -            & 48            & 48     & 48             & 48 & -             & -  & -  & - \\
tau                  & -             & -             & 12             & 12      & 12            & 12 & -             & -  & - & - \\
\bottomrule
\end{tabular}
\end{table}

\begin{table}[ht]
    \centering
    \tiny
    \begin{minipage}{0.49\textwidth}
        \centering
\caption{Experiments hyperparameters for GPT-2 medium pretraining experiments}
\label{tab:hyperparams-pretraining-appendix}
\begin{tabular}{@{}llllll@{}}
\toprule
                     & DDP           & CO2           & SlowMo        & LayUp / GoSGD  \\ \midrule
batch\_size          & 384            & 384            & 384            & 384      \\
optimizer            & AdamW         & AdamW         & AdamW         & AdamW  \\
lr                   & 0.76978       & 0.9645        & 0.9645        & 0.00075249\\
Cosine scheduler (T\_max)               & 45,539         & 45,539         & 45539         & 138,972 \\
warm\_up\_iterations & 0             & 0             & 0             & 3488  \\
warm\_up lr          & 0             & 0             & 0       	     & 0.00014011 \\
weight\_decay        & 0             & 0             & 0             & 0.0031235      \\
out\_freq            & -             & 20            & 20            & -      \\
tau                  & -             & 5             & 5             & -      \\
\bottomrule
\end{tabular}
    \end{minipage}
    \hfill
    \begin{minipage}{0.49\textwidth}
        \centering
\caption{Experiments hyperparameters for GPT-2 XL fine-tuning experiments}
\label{tab:hyperparams-fine-tuning-appendix}
\begin{tabular}{@{}llllll@{}}
\toprule
                     & DDP           & CO2           & SlowMo        & LayUp / GoSGD  \\ \midrule
batch\_size          & 16             & 16             & 16             & 16     \\
optimizer            & AdamW         & AdamW         & AdamW         & AdamW  \\
lr                   & 0.31645       & 0.38939       & 0.32855       & 0.0003 \\
Cosine scheduler (T\_max)               & 7,286         & 7,286         & 7,286         & 29,144 \\
warm\_up\_steps & 0             & 0             & 0             & 7,000  \\
warm\_up lr          & 0.005         & 0.005         & 0.005      	 & 0.0002 \\
weight\_decay        & 0             & 0             & 0             & 0      \\
out\_freq            & -             & 48            & 48            & -      \\
tau                  & -             & 12            & 12            & -      \\
\bottomrule
\end{tabular}
    \end{minipage}
\end{table}

\begin{table}[h]
\centering
\caption{Experiments hyperparameters for ImDB}
\label{tab:hyperparams-imdb-appendix}
\begin{tabular}{@{}llllll@{}}
\toprule
                     & DDP               & LayUp  \\ \midrule
batch\_size          & 75                 & 75      \\
optimizer            & Adam             & Adam  \\
lr                   & 0.001           & 0.0015 \\
Cosine scheduler (T\_max)   & 150         & 150 \\
warm\_up\_epochs & 0                 & 0  \\
warm\_up lr          & 0         	 & 0 \\
weight\_decay        & 0                 & 0      \\
\bottomrule
\end{tabular}
\end{table}
\section{Implementation details.}
\label{appendix-implementation}

All algorithms were implemented using LibTorch, the C++ front-end of the Torch library.
For the ResNet architectures, we use the code from~\citet{resnet_cxx} and the GPT-2 architectures were implemented based on~\citet{nanoGpT}. We submit the code as supplementary material.

\subsection{Baseline implementation}
DDP was implemented based on LibTorch distributed training example~\cite{pytorch} combined with NCCL~\cite{nccl} for efficient averaging, MPI~\citep{open_mpi} for multiprocessing, OpenMP~\cite{openmp} for multithreading. 
SlowMo is a simple extension of the DDP code, which performs global averaging and outer momentum steps at fixed iteration interval. For CO2, the global averaging step is performed inside a thread to allow computation communication overlap.
GoSGD code is adapted from LayUp implementation.

\subsection{LayUp implementation}

\paragraph{Computation thread.} 
LayUp was implemented in a single root process. 
We adopted a cascade approach where we split the devices into three (3) groups, called in the following $G_0$, $G_1$ and $G_2$. Only devices within the same group are running the forward pass at the same time.

When $G_1$ is performing its forward passes, once it finishes, it starts with the corresponding backward passes and at the same time $G_2$ starts with its forward passes. The updater thread attached to $G_1$ will then send in parallel send updates, layer-wise, at random 
to any GPU. The choice of the receiving device is done at the beginning of each iteration. The case where devices of a group can only send updates to devices pertaining to another group has been explored but no significant differences was noticed.

\textbf{Updater thread.} Each device has an updater thread attached to it and it has two roles: check if gradients of a layer have already been computed during the backward pass and to carry out communication with the push-sum protocol using NCCL. 
Although NCCL primitives are non-blocking, they are also not thread-safe making its usage incompatible with fully asynchronous communication. 
To solve the problem, we ensure that all NCCL calls are sequential using a lock. 
Note that the lock is only active during the call of the primitive (initiation of the call) but not its execution because the execution itself is non-blocking.
Therefore, the communication overlaps with the computation. 
Moreover, multiple updater threads can update the same parameters simultaneously (lock-free) leading to the updates being overwritten.
Empirically, this doesn't affect convergence significantly.

\section{Complete Proofs}
\label{appendix:proofs}

In this section, we provide proofs for the various convergence results from the main text. We consider the different weights $w_i$ to be stochastic, time-varying but don't add time subscript to simplify the notation.

\subsection{Further results}

We also provide convergence results when assuming the objective function to be strongly-convex 
and showing convergence of the consensus iterates to the true parameter:

\begin{theorem}
    \label{convex_theorem}
    Considering the SGD iterations of algorithm \ref{alg:async} and satisfying assumptions \ref{bias_equation}, \ref{variance_bound}, \ref{strong_convexity}, \ref{elastic_consistencey}. Given a smooth, strongly-convex objective function $f$, with unknown minimum at $x^*$ and learning rate $\eta = \frac{2(\log T + \log M)}{cT}$,  where $T \ge \frac{576K_f^2M}{c^2}$ is the number of iterations.
\begin{align*}
\EX \norm{\bar{x}_{T} - x^*}^2 &= \bigO\left(\frac{ \norm{\bar{x}_0 - x^*}^2}{TM} + \frac{8(\log T + \log M)\sigma^2}{c^2TM} + \frac{240(\log T + \log M)^3 B^{'2}\tau_{max}^2K^2}{c^4T^3} + \frac{160(\log T + \log M)^2 B^{'2}\tau_{max}^2K^2}{c^4T^2} \right) \notag
        \label{convex_theorm}
    \end{align*}
\end{theorem}

\subsection{Assumptions on gradients and objective function}
\label{appendix:assumption}
We adapt common assumptions for SGD convergence analysis:
\begin{enumerate}
    \item \textbf{Biased stochastic gradients.} The bias $b$ induced by layer-wise updates can be expressed as:
    \label{bias_equation}
    \begin{align}
       \forall x \in \mathbb{R}^d, \EX \left[g(x)\right] = \nabla f(x) + b(x).
    \end{align}
    \item \textbf{Bounded variance.} 
    \label{variance_bound}
    \begin{align}
        \forall x \in \mathbb{R}^d, \EX \left[\norm{g(x) - \nabla f(x) - b(x)}^2\right] \leq \sigma^2.
    \end{align}
    \item \textbf{Bounded second moment.}
    \label{gradient_bound}
    \begin{align}
        \forall x \in \mathbb{R}^d, \EX \left[\norm{g(x)}^2\right] \leq S^2.
    \end{align}
    \item \textbf{Smoothness.} The function $f$: $\mathbb{R}^d \to \mathbb{R}$ is smooth iff:
    \begin{align}
        \norm{\nabla f(x) - \nabla f(y)} \leq K_f\norm{x - y} \text{ for } K_f > 0.
    \end{align}
    The stochastic biased gradient $g$ is $K_b$-Lipschitz:
    \begin{align}
        \norm{g(x) - g(y)} \leq K_b\norm{x - y} \text{ for } K_b > 0.
    \end{align}
    \label{smoothness}
    \item \textbf{Strong convexity.} 
    \label{strong_convexity}
    The objective function $f$ is c-strongly convex  for $x, y \in \mathbb{R}^d$ and $\forall c > 0$:
    \begin{align}
        (x - y)^T(\nabla f(x) - \nabla f(y)) \geq c\norm{x - y}^2.
    \end{align}
    \item \textbf{Elastic Consistency (E.C) bound.} 
    \label{elastic_consistencey}
    We adapt the elastic consistency bound to model the convergence of the consensus.
    \begin{align}
        \EX \left[\norm{\bar{x}_t - x_t^i}^2 \right] \leq \eta^2B^2
    \end{align}
    with $B = B'\tau_{max}$ and $B' =  \frac{(M-1)S}{M}$ (Lemma 15 in~\citet{nadiradze_2021_elastic}),  $\tau_{max}$ the maximum communication delay allowed and $\eta$ the learning rate.
\end{enumerate}

\subsection{Useful inequalities}

\todo{I would find it nicer to give the definitions directly incline where used the first time}
\begin{enumerate}

    \item \textbf{Cauchy-Schwarz}. Given a set of n vectors \{$a_i \in \mathbb{R}^d$\}, $1 \leq i \leq n$ :
        $$ \norm{\sum_{i=1}^n a_i}^2 \leq n \sum_{i=1}^n \norm{a_i^2}$$

    \item \textbf{Young's inequality}. Given two vectors $a, b \in \mathbb{R}^d$ and $\gamma > 0$:
        $$ 2 \langle a, b \rangle \leq \gamma \norm{a}^2 + \gamma^{-1} \norm{b}^2$$

    \item \textbf{Lemma 9 from~\cite{nadiradze_2021_elastic}}. Given $f$ is an L-smooth, c-strongly convex, with unknown minimum $x^*$ and any vector $x$ :
    $$ \langle \nabla f(x), x - x^* \rangle \ge \frac{1}{2L} \norm{\nabla f(x)}^2 + \frac{c}{2} \norm{x -x ^*}^2$$

\end{enumerate}

\subsection{Proof of gradients bias bound}
Here we derive the bound of the second moment of the bias, $b$, induced by the algorithm as a difference between true gradients and used gradients during the backward pass.

\textbf{Lemma \ref{bias_bound}.}
Under Assumptions \ref{smoothness} and \ref{elastic_consistencey}, the second moment of the gradients bias induced by algorithm \ref{alg:async} is bounded.
\begin{equation}
    \forall x \in \mathbb{R}^d, \EX \left[ \norm{b(x)}^2\right] \leq 4K_b^2\eta^2B^2  \,.
\end{equation}

with $B = B'\tau_{max}$

\begin{proof}
The gradients bias is expressed as the difference between the derivative evaluated at the point where the forward pass was executed, $x^i$ and derivative evaluated at a different point, $x^j$, than forward pass. Using the Lipschitz-continuity assumption, we define the bias $b(x^i)$ at device $i$, in short $b_i$ as: 
    \begin{align*}
        \EX \left[ \norm{b(x^i)}^2 \right] = \EX \left[ \norm{g(x^i) - g(x^j)}^2\right] &\leq K_b^2 \EX \left[\norm{x^i - x^j}^2\right] \\
        &= K_b^2\EX \left[\norm{x^i - \bar{x} + \bar{x} - x^j}^2\right] \\
&\stackrel{\textit{Cauchy-schwarz}}{\leq} 2K_b^2 (\EX \left[\norm{\bar{x} - x^i}^2 \right] + \EX \left[\norm{\bar{x} - x^j}^2 \right]) \\
        &\stackrel{\textit{\ref{elastic_consistencey}}}{\leq} 4K_b^2\eta^2B^2
    \end{align*}
\end{proof}

\subsection{Convergence in non-convex case}

To establish the convergence of LayUp when the objective function, we first establish an intermediate result:
\begin{lemma}
    \label{non_convex_lemma}
    Considering SGD iterations form algorithm \ref{alg:async}, for a smooth non-convex objective function $f$ and learning rate $\eta \leq \frac{1}{12K_f}$ we have:
\end{lemma}

$$\EX \left[ f(\bar{x}_{t+1})\right] \leq \EX  f(\bar{x}_t) - \frac{\eta}{4} \EX \norm{\nabla f (\bar{x}_t)}^2+ 15\eta^4K^3B^2+ \frac{K \eta^2 \sigma^2}{M} + \frac{5 \eta^3K^2B^2}{2}$$

\todo{We would need a bit of text in this long proof so it is easier to follow.}
\begin{proof}
    We follow the steps from \citet{nadiradze_2021_elastic} and condition on $\bar{x}_t$ and all $x_t^i$ and compute the expectation $\EX_s$. Using the descent lemma with biased gradients, we have, by taking into account the contribution of each device:
    \begin{align*}
        \EX_S \left[ f(\bar{x}_{t+1})\right] &\leq f(\bar{x}_t) - \eta\sum_{i=1}^M  \EX_S \left[ \langle w_ig(x_t^i), \nabla f(\bar{x}_t) \rangle\right] + \frac{K_f \eta^2}{2} \EX_s \norm{\sum_{i=1}^Mw_ig(x_t^i)}^2 \\
        &\stackrel{\textit{bias equation}}{=} f(\bar{x}_t) - \eta\norm{\nabla f(\bar{x}_t)}^2 + \frac{\eta}{M} \sum_{i=1}^M \langle \nabla f(\bar{x}_t) -\nabla f(x_t^i), \nabla f(\bar{x}_t) \rangle
        + \frac{\eta}{M} \sum_{i=1}^M \langle -²b_i, \nabla f(\bar{x}_t) \rangle\\ &+ \frac{K_f\eta^2}{2} \EX_S \norm{\sum_{i=1}^M w_i(g(x_t^i) - \nabla f(x_t^i) - b_i + \nabla f(x_t^i) - \nabla f(\bar{x}_t) + b_i + \nabla f(\bar{x}_t))}^2 \\
        &\stackrel{\textit{Cauchy-Schwarz}}{\leq} f(\bar{x}_t) - \eta \norm{\nabla f(\bar{x}_t)}^2 + \frac{\eta}{M} \sum_{i=1}^M \langle \nabla f(\bar{x}_t)-\nabla f(x_t^i), \nabla f(\bar{x}_t) \rangle \\
        &+ \frac{\eta}{M} \sum_{i=1}^M \langle -b_i, \nabla f(\bar{x}_t) \rangle + K_f\eta^2  \sum_{i=1}^M \EX_S \norm{w_i(g(x_t^i) - \nabla f(x_t^i) - b_i)}^2 \\
        &+ K_f\eta^2  \sum_{i=1}^M \EX_S \norm{w_i(\nabla f(x_t^i) -\nabla f(\bar{x}_t) + b_i + \nabla f(\bar{x}_t))}^2 \\
        &\stackrel{\textit{Cauchy-Schwarz}}\leq f(\bar{x}_t) - \eta \norm{\nabla f(\bar{x}_t)}^2 + \frac{\eta}{M} \sum_{i=1}^M \langle \nabla f(\bar{x}_t) - \nabla f(x_t^i), \nabla f(\bar{x}_t) \rangle + \frac{\eta}{M} \sum_{i=1}^M \langle -b_i, \nabla f(\bar{x}_t) \rangle \\ 
        &+ \frac{K_f \eta^2}{M} \sigma^2 + \frac{3MK_f\eta^2}{M^2}(  \sum_{i=1}^M \norm{\nabla f(x_t^i) - \nabla f(\bar{x}_t)}^2 + \sum_{i=1}^M \norm{b_i}^2 + \sum_{i=1}^M \norm{\nabla f(\bar{x}_t)}^2) \\
        &\stackrel{\textit{Young}}{\leq}f(\bar{x}_t) - \eta \norm{\nabla f (\bar{x}_t)}^2 + \frac{\eta}{M} \sum_{i=1}^M \norm{\nabla f ( \bar{x}_t) - \nabla f(x_t^i)}^2 + \frac{\eta}{4} \norm{\nabla f(\bar{x}_t})^2 + \frac{\eta}{4} \norm{\nabla f(\bar{x}_t)}^2 \\
        &+ \frac{\eta}{M} \sum_{i=1}^M \norm{b_i}^2 + \frac{3 K_f \eta^2}{M}   \sum_{i=1}^M \norm{\nabla f(x_t^i) - \nabla f(\bar{x}_t)}^2 + \sum_{i=1}^M \norm{b_i}^2 + \sum_{i=1}^M \norm{\nabla f (\bar{x}_t}^2 + \frac{K_f \eta^2 \sigma^2}{M} \\
        &\stackrel{\textit{smoothness}}{\leq} f(\bar{x}_t) - \frac{\eta}{2} \norm{\nabla f (\bar{x}_t)}^2 + \frac{\eta K_f^2}{M} \sum_{i=1}^M \norm{\bar{x}_t - x_t^i}^2 + \frac{\eta}{M} \sum_{i=1}^M \norm{b_i}^2 + \frac{3\eta^2 K_f^3}{M} \sum_{i=1}^M \norm{x_t^i - \bar{x}_t}^2 \\
        &+  \frac{3\eta^2 K_f}{M} \sum_{i=1}^M \norm{b_i}^2 + 3K_f\eta^2 \norm{\nabla f(\bar{x}_t)}^2 + \frac{K_f \eta^2 \sigma^2}{M}
\end{align*}
\begin{align*}
\EX \left[ f(\bar{x}_{t+1})\right] = \EX\left[\EX\left[ f(\bar{x}_{t+1} \mid \bar{x}_t, x_t^i\right]\right] \\
&= \EX \left[ f(\bar{x}_t)\right] - \frac{\eta}{2} \EX \left[\norm{\nabla f (\bar{x}_t)}^2\right] + \frac{\eta K_f^2}{M} \sum_{i=1}^M \EX \left[\norm{\bar{x}_t - x_t^i}^2 \right] + \frac{\eta}{M} \sum_{i=1}^M \EX \left[\norm{b_i}^2\right] \\ 
&+ \frac{3\eta^2 K_f^3}{M} \sum_{i=1}^M \EX \left[\norm{x_t^i - \bar{x}_t}^2 \right] 
+ \frac{3\eta^2 K_f}{M} \sum_{i=1}^M \EX \left[\norm{b_i}^2 \right] + 3K_f\eta^2 \EX \left[ \norm{\nabla f(\bar{x}_t)}^2 \right] + \frac{K_f \eta^2 \sigma^2}{M} \\
&\stackrel{\textit{bias-bound and E.C.}}{\leq} \EX \left[ f(\bar{x}_t)\right] - \frac{\eta}{2} \EX \left[\norm{\nabla f (\bar{x}_t)}^2\right] + \eta^3 B^2 K_f^2  + 4\eta^3B^2K_b^2 + 3\eta^4B^2K_f^3 \\
&+ 12\eta^4B^2K_fK_b^2 + 3K_f\eta^2 \EX \left[ \norm{\nabla f(\bar{x}_t)}^2 \right] + \frac{K_f \eta^2 \sigma^2}{M}
\end{align*}
\begin{align*}
\text{choosing such that: }\eta \leq \frac{1}{12K_f} \text{ and } K = \max \{K_f, K_b\}\\
\EX \left[ f(\bar{x}_{t+1})\right] &\leq \EX \left[ f(\bar{x}_t)\right] - \frac{\eta}{4} \EX \norm{\nabla f (\bar{x}_t)}^2+ 15\eta^4B^2K^3+ \frac{K \eta^2 \sigma^2}{M} + 5\eta^3B^2K^2
     \end{align*}
\end{proof}

\textbf{Theorem \ref{non_convex_theorem}}

Considering the SGD iterations of algorithm \ref{alg:async} and satisfying assumptions \ref{bias_equation}, \ref{variance_bound}, \ref{elastic_consistencey}. Given a smooth, non-convex objective function $f$, with unknown minimum at $x^*$ and learning rate $\eta = \frac{\sqrt{M}}{\sqrt{T}}$, where $T \ge 144K_f^2M$ is the number of iterations.

    \begin{align}
        \frac{1}{T} \sum_{t=0}^{T-1} \EX \left[\norm{\nabla f (\bar{x}_t)}^2)\right] \leq \frac{4 (f(\bar{x}_0) - f(x^*))}{\sqrt{TM}}  + \frac{60 K^3 B^2 M\sqrt{M}}{T\sqrt{T}}+ \frac{4K\sigma^2}{\sqrt{TM}} + \frac{20K^2B^2M}{T} \notag
    \end{align}

\begin{proof}
    By Lemma \ref{non_convex_lemma}, we have:
    $$\EX \left[ f(\bar{x}_{t+1})\right] \leq \EX \left[ f(\bar{x}_t)\right] - \frac{\eta}{4} \EX \norm{\nabla f (\bar{x}_t)}^2+ 15\eta^4K^3B^2+ \frac{K \eta^2 \sigma^2}{M} + 5 \eta^3K^2B^2$$

Summing over t $\in$ [0, T-1], we get:
\begin{align*}
    \sum_{t=0}^{T-1} \EX \left[ f(\bar{x}_{t+1})\right] \leq \sum_{t=0}^{T-1} \EX \left[ f(\bar{x}_t)\right]  - \frac{\eta}{4} \EX \left[\norm{\nabla f (\bar{x}_t)}^2\right] + 15\eta^4K^3B^2+ \frac{K \eta^2 \sigma^2}{M} + 5 \eta^3K^2B^2
\end{align*}
Rearranging the inequality:
\begin{align*}
    \sum_{t=0}^{T-1} \frac{\eta}{4} \EX \left[\norm{\nabla f (\bar{x}_t)}^2\right] \leq (f(\bar{x}_0) - \EX\left[f(\bar{x}_T)\right]) + 15\eta^4K^3B^2 T+ \frac{K \eta^2 \sigma^2 T}{M} + 5 \eta^3K^2B^2 T
\end{align*}
Dividing by $\frac{\eta T}{4}$:
\begin{align*}
    \frac{1}{T} \sum_{t=0}^{T-1} \EX \left[\norm{\nabla f (\bar{x}_t)}^2\right] \stackrel{\textit{$x*$ is optimum}}{\leq} \frac{4 (f(\bar{x}_0) - f(x^*))}{\eta T} + 60 \eta^3 K^3 B^2 + \frac{4K\eta\sigma^2}{M} + 20\eta^2K^2B^2
\end{align*}
Replacing $\eta$ by its value:
\begin{align*}
    \frac{1}{T} \sum_{t=0}^{T-1} \EX \left[\norm{\nabla f (\bar{x}_t)}^2\right] \leq \frac{4 (f(\bar{x}_0) - f(x^*))}{\sqrt{TM}} + \frac{60 K^3 B^2 M\sqrt{M}}{T\sqrt{T}}+ \frac{4K\sigma^2}{\sqrt{TM}} + \frac{20K^2B^2M}{T}   
\end{align*}
\end{proof}

\subsection{Convergence in strongly convex case}
\begin{lemma}
\label{convex_lemma}
Consider SGD iterations defined in algorithm \ref{alg:async} and satisfying elastic consistency bound \ref{elastic_consistencey}. For a smooth,
strongly convex objective function f and the constant learning rate $\eta \leq \frac{1}{6K_f}$. We have that:
$$\EX \left[\norm{\bar{x}_{t+1} - x^*}^2\right] \leq (1-\frac{\eta c}{2}) \norm{\bar{x}_t - x^*}^2 + 30\eta^4B^2K^2 + \frac{20\eta^3B^2K^2}{c} + \frac{2\eta^2\sigma^2}{M} $$
\end{lemma}

\begin{proof}
    We follow the steps from \citet{nadiradze_2021_elastic} and condition on $\bar{x}_t$ and all $x_t^i$ and compute the expectation $\EX_S$. Using the descent lemma with biased gradients, we have:

    \begin{align*}
         \EX_S \left[\norm{\bar{x}_{t+1} - x^*}^2\right] &= \EX_S \left[\norm{\bar{x}_t - \eta\sum_{i=1}^M w_ig(x_t^i) - x^*}^2\right] \\
         &=\norm{\bar{x}_t - x^*}^2 - 2\eta\sum_{i=1}^M \EX_S\left[ \langle w_ig(x_t^i), \bar{x}_t - x^* \rangle \right] + \EX_S \norm{\sum_{i=1}^M w_ig(x_t^i)}^2 \\
         &\stackrel{\textit{bias equation}}{=} \norm{\bar{x}_t - x^*}^2 - 2\eta \langle \nabla f(\bar{x}_t), \bar{x}_t - x^* \rangle + \frac{2\eta}{M}\sum_{i=1}^M \langle \nabla f(\bar{x}_t) - \nabla f(x_t^i), \bar{x}_t - x^* \rangle 
         \\
         &\quad + \frac{2\eta}{M}\sum_{i=1}^M \langle -b_i, \bar{x}_t - x^* \rangle + \eta^2 \EX_S\norm{\sum_{i=1}^M w_i(g(x_t^i) - \nabla f(x_t^i) -b_i + \nabla f(x_t^i) - \nabla f(\bar{x}_t) + \nabla f(\bar{x}_t) + b_i)}^2 \\
         &\stackrel{\textit{Cauchy-Schwarz}}{\leq} \norm{\bar{x}_t - x^*}^2 - 2\eta \langle \nabla f(\bar{x}_t), \bar{x}_t - x^* \rangle + \frac{2\eta}{M}\sum_{i=1}^M \langle \nabla f(\bar{x}_t) - \nabla f(x_t^i), \bar{x}_t - x^* \rangle \\
         &\quad+ \frac{2\eta^2}{M^2} \sum_{i=1}^M\EX_S\norm{g(x_t^i) - \nabla f(x_t^i) -b_i}^2 
         + \sum_{i=1}^M\frac{6\eta^2}{M}\norm{\nabla f(x_t^i) - \nabla f(\bar{x}_t)}^2  \\
         &\quad + \frac{6\eta^2}{M}\sum_{i=1}^M \norm{\nabla f(\bar{x}_t)}^2
         + \frac{6\eta^2}{M} \sum_{i=1}^M\norm{b_i}^2 \\
         &\stackrel{\textit{bounded-variance}}{\leq} \norm{\bar{x}_t - x^*}^2 - 2\eta \langle \nabla f(\bar{x}_t), \bar{x}_t - x^* \rangle + \frac{2\eta}{M}\sum_{i=1}^M \langle \nabla f(\bar{x}_t) - \nabla f(x_t^i), \bar{x}_t - x^* \rangle \\
         &+ \frac{2\eta}{M}\sum_{i=1}^M \langle -b_i, \bar{x}_t - x^* \rangle
         + \frac{2\eta^2\sigma^2}{M^2}
         + \frac{6\eta^2}{M}\sum_{i=1}^M \norm{\nabla f(x_t^i) - \nabla f(\bar{x}_t)}^2 \\
         &+ \frac{6\eta^2}{M}\sum_{i=1}^M\norm{\nabla f(\bar{x}_t)}^2 
         + \frac{6\eta^2}{M}\sum_{i=1}^M\norm{b_i}^2 \\
         &\stackrel{\textit{Lemma 9 from E.C}}{\leq} \norm{\bar{x}_t - x^*}^2 - \frac{\eta}{K} \norm{\nabla f(\bar{x}_t)}^2 - \eta c \norm{\bar{x}_t - x^*}^2 + \frac{2\eta}{M}\sum_{i=1}^M \langle \nabla f(\bar{x}_t) - \nabla f(x_t^i), \bar{x}_t - x^* \rangle \\
         &+ \frac{2\eta}{M}\sum_{i=1}^M \langle -b_i, \bar{x}_t - x^* \rangle 
         + \frac{6\eta^2}{M}\sum_{i=1}^M\norm{\nabla f(x_t^i) - \nabla f(\bar{x}_t)}^2 \\
         &+ \frac{6\eta^2}{M}\sum_{i=1}^M\norm{\nabla f(\bar{x}_t)}^2
         + \frac{6\eta^2}{M}\sum_{i=1}^M\norm{b_i}^2
         + \frac{2\eta^2\sigma^2}{M^2}\\
        &\stackrel{\textit{Young}}{\leq} \norm{\bar{x}_t - x^*}^2 - \frac{\eta}{K} \norm{\nabla f(\bar{x}_t)}^2 - \eta c \norm{\bar{x}_t - x^*}^2 
        + \frac{4\eta}{cM}\sum_{i=1}^M \norm{f(\bar{x}_t) - \nabla f(x_t^i)}^2
        + \frac{c\eta}{4M} \sum_{i=1}^M \norm{\bar{x}_t - x^*}^2 \\
        &\quad + \frac{c\eta}{4M} \sum_{i=1}^M \norm{\bar{x}_t - x^*}^2  + \frac{4\eta}{cM} \sum_{i=1}^M \norm{b_i}^2 + \frac{6\eta^2}{M}\sum_{i=1}^M \norm{\nabla f(x_t^i) - \nabla f(\bar{x}_t)}^2 \\
        &\quad 
        + \frac{6\eta^2}{M}\sum_{i=1}^M\norm{\nabla f(\bar{x}_t)}^2
         + \frac{6\eta^2}{M}\sum_{i=1}^M\norm{b_i}^2
         + \frac{2\eta^2\sigma^2}{M^2}\\
         &\stackrel{\textit{Smoothness}}{\leq} (1-\frac{\eta c}{2}) \norm{\bar{x}_t - x^*}^2 - (\frac{\eta}{K} - 6\eta^2)\norm{\nabla f(\bar{x}_t)}^2 + \frac{4\eta K_f^2}{C M}\sum_{i=1}^M\norm{\bar{x}_t - x_t^i}^2
         + \frac{4\eta}{cM} \sum_{i=1}^M \norm{b_i}^2\\
        &\quad + \frac{6\eta^2K_f^2}{M} \sum_{i=1}^M\norm{x_t^i -  \bar{x}_t}^2
         + \frac{6\eta^2}{M}\sum_{i=1}^M\norm{b_i}^2  
         + \frac{2\eta^2\sigma^2}{M^2}\\
         \text{choosing such that: }\eta \leq \frac{1}{6K_f} \\
         &\leq (1-\frac{\eta c}{2}) \norm{\bar{x}_t - x^*}^2 + \frac{4\eta K}{c M}\sum_{i=1}^M \norm{\bar{x}_t - x_t^i}^2 
         + \frac{4\eta}{cM} \sum_{i=1}^M \norm{b_i}^2 \\
        &\quad + \frac{6\eta^2K_f^2}{M} \sum_{i=1}^M\norm{x_t^i - \bar{x}_t}^2
         + \frac{6\eta^2}{M}\sum_{i=1}^M\norm{b_i}^2
         + \frac{2\eta^2\sigma^2}{M^2}\\
\end{align*}
\begin{align*}
\EX \left[\norm{\bar{x}_{t+1} - x^*}^2\right] &= \EX \left[ \EX \left[ \norm{\bar{x}_{t+1} - x^* \mid \bar{x}_t, x_t^i}^2 \right] \right] \\
&= (1-\frac{\eta c}{2}) \norm{\bar{x}_t - x^*}^2 + \frac{4\eta K_f^2}{c M}\sum_{i=1}^M \EX\norm{\bar{x}_t - x_t^i}^2 + \frac{6\eta^2K_f^2}{M} \sum_{i=1}^M\EX \norm{x_t^i - \bar{x}_t}^2  \\
&\quad+ \frac{6\eta^2}{M} \sum_{i=1}^M\EX \norm{b_i}^2 + \frac{4\eta}{cM}   \sum_{i=1}^M\EX\norm{b_i}^2
+ \frac{2\eta^2\sigma^2}{M^2}\\
&\stackrel{\textit{E.C and bias-bound}}{\leq} (1-\frac{\eta c}{2}) \norm{\bar{x}_t - x^*}^2 + \frac{4\eta^3B^2K_f^2}{c} + 6\eta^4B^2K_f^2 + 24\eta^4K_b^2B^2 + \frac{16\eta^3B^2K_b^2}{c} + \frac{2\eta^2\sigma^2}{M} \\
&\leq (1-\frac{\eta c}{2}) \norm{\bar{x}_t - x^*}^2 + 30\eta^4B^2K^2 + \frac{20\eta^3B^2K^2}{c} + \frac{2\eta^2\sigma^2}{M} 
    \end{align*}
\end{proof}

\textbf{Theorem \ref{convex_theorem}}
Considering the SGD iterations of algorithm \ref{alg:async} and satisfying assumptions \ref{bias_equation}, \ref{variance_bound}, \ref{strong_convexity}, \ref{elastic_consistencey}. Given a smooth, strongly-convex objective function $f$, with unknown minimum at $x^*$ and learning rate $\eta = \frac{2(\log T + \log M)}{cT}$, where $T \ge 576K_f^2M$ is the number of iterations.

$$\EX \left[\norm{\bar{x}_{T} - x^*}^2\right] \leq \frac{ \norm{\bar{x}_0 - x^*}^2}{TM} + \frac{160(\log T + \log M)^2 B^2K^2}{c^4T^2} + \frac{240(\log T + \log M)^3 B^2K^2}{c^4T^3} + \frac{8(\log T + \log M)\sigma^2}{c^2TM}$$

\begin{proof}
By Lemma \ref{convex_lemma}, we have:
$$\EX \left[\norm{\bar{x}_{t+1} - x^*}^2\right] = (1-\frac{\eta c}{2}) \norm{\bar{x}_t - x^*}^2 + 30\eta^4B^2K^2 + \frac{20\eta^3B^2K^2}{c} + \frac{2\eta^2\sigma^2}{M} $$

Summing over t $\in$ [0, T-1]
\begin{align*}
    \EX \left[\norm{\bar{x}_{T} - x^*}^2\right] &\leq (1-\frac{\eta c}{2}) \norm{\bar{x}_0 - x^*}^2 + \sum_{t=0}^{T-1} (1-\frac{\eta c}{2})^t (30\eta^4B^2K^2 + \frac{20\eta^3B^2K^2}{c} + \frac{2\eta^2\sigma^2}{M}) \\
    &\leq (1-\frac{\eta c}{2}) \norm{\bar{x}_0 - x^*}^2 + \sum_{t=0}^{\infty} (1-\frac{\eta c}{2})^t (30\eta^4B^2K^2 + \frac{20\eta^3B^2K^2}{c} + \frac{2\eta^2\sigma^2}{M}) \\
    &\leq e^{-\frac{\eta c T}{2}} \norm{\bar{x}_0 - x^*}^2 + \frac{40\eta^2B^2K^2}{c^2} + \frac{60\eta^3B^2K^2}{c} + \frac{4\eta\sigma^2}{cM}\\
    &= \frac{ \norm{\bar{x}_0 - x^*}^2}{TM} + \frac{160(\log T + \log M)^2 B^2K^2}{c^4T^2} + \frac{240(\log T + \log M)^3 B^2K^2}{c^4T^3} + \frac{8(\log T + \log M)\sigma^2}{c^2TM}
\end{align*}
\end{proof}




\end{document}